\documentclass[11pt]{article}

\usepackage[final]{acl}

\usepackage{times}
\usepackage{latexsym}
\usepackage{tabularx}
\usepackage{multirow}
\usepackage{amssymb}
\usepackage{array}
\newcolumntype{Y}{>{\centering\arraybackslash}X}
\usepackage{booktabs}
\usepackage{array}
\usepackage{graphicx}
\usepackage{pifont}
\usepackage[table]{xcolor}

\newcommand{\cmark}{\ding{51}}
\newcommand{\rot}[1]{\rotatebox{90}{#1}}

\usepackage[T1]{fontenc}

\usepackage[utf8]{inputenc}

\usepackage{microtype}

\usepackage{inconsolata}

\usepackage{graphicx}

\usepackage{times}
\usepackage{latexsym}
\usepackage[T1]{fontenc}
\usepackage[utf8]{inputenc}
\usepackage{microtype}
\usepackage{inconsolata}
\usepackage{graphicx}
\usepackage{enumitem}
\usepackage{booktabs}
\usepackage{amsmath}
\usepackage{xcolor}
\usepackage[most]{tcolorbox}
\usepackage[textsize=footnotesize]{todonotes}

\providecommand{\cmark}{\ding{51}}

\title{ImageEval 2026: Culturally Grounded Arabic Multimodal Evaluation}

\author{
Samir Abdaljalil$^1$\thanks{Equal contribution.},
Hunzalah Hassan Bhatti$^2$\footnotemark[1],
Ahlam Bashiti$^3$,
Farina Amir$^4$,\\
\textbf{Md Arid Hasan$^5$,
Basel Mousi$^2$,
Nadir Durrani$^2$,
Fahim Dalvi$^2$,}\\
\textbf{Zien Sheikh Ali$^2$,
Erchin Serpedin$^1$,
Hasan Kurban$^4$,
Mustafa Jarrar$^4$,}\\
\textbf{Shammur Absar Chowdhury$^2$,
Firoj Alam$^2$}\\[2mm]
$^1$Texas A\&M University, 
$^2$Qatar Computing Research Institute, Qatar\\
$^3$Birzeit University, Palestine, 
$^4$Hamad Bin Khalifa University, Qatar\\
$^5$University of Toronto, Canada\\
\texttt{\url{https://imageeval2026.github.io/}}
}

\begin{document}
\maketitle

\begin{abstract}
We present an overview of the \textbf{ImageEval 2026} shared task on culturally grounded Arabic multimodal evaluation. It includes two tasks: ({\em i})~\textbf{AynVQA}, covering spoken visual question answering and image-grounded hallucination detection in English and Modern Standard Arabic (MSA), and ({\em ii})~\textbf{CRAI-Bench}, evaluating the cultural accuracy of text-to-image generation. A total of 14 teams participated in the test phase, with 12 teams submitting system description papers. Participating systems used a range of approaches, including zero-shot prompting, fine-tuning of vision-language models, speech-recognition pipelines, ensembling, and score calibration. We describe the task setup, datasets, evaluation procedure, and participating systems, and summarize the main results across the different tracks. All datasets and evaluation scripts from the shared task are released to the research community.
The shared task highlights the challenges of culturally grounded multimodal evaluation, particularly for Arabic speech and image-text reasoning.
\end{abstract}

\section{Introduction}

Recent benchmarking efforts have evaluated the capabilities of LLMs, speech-language models, and vision-language models (VLMs) across tasks such as spoken question answering (SQA) and visual question answering (VQA)~\citep{hudson2019gqa,balanced_vqa_v2}. These benchmarks have driven substantial progress in multimodal reasoning and compositional understanding. However, most focus on general scene understanding and provide limited evidence of models' ability to reason about culturally specific concepts or remain robust across languages and regional varieties. Cross-lingual benchmarks such as xGQA extend VQA beyond English, yet largely preserve the same underlying visual domain~\citep{pfeiffer-etal-2022-xgqa}.

To address this limitation, culture-centered benchmarks such as CulturalVQA~\citep{nayak-etal-2024-benchmarking}, CVQA~\citep{romero2024cvqa}, SEA-VQA~\citep{urailertprasert-etal-2024-sea}, and OASIS~\citep{alam2025everydaymmqa} broaden evaluation to culturally situated concepts, including artifacts, food, clothing, practices, landmarks, and regional identities. Results on these benchmarks show that VLMs continue to struggle with culturally grounded understanding, highlighting an important distinction: multilingual coverage does not necessarily translate into cultural understanding.

These challenges are also closely related to multimodal hallucination, where models generate plausible responses that are not sufficiently grounded in the visual input~\citep{chen2026survey}. In culturally situated settings, a model may rely on linguistic or cultural associations to infer an answer even when the image provides insufficient evidence. Distinguishing genuine visual understanding from such associative reasoning becomes particularly difficult when incorrect alternatives are themselves culturally plausible. Consequently, answer accuracy alone may not capture visual grounding, and evaluation should assess whether models distinguish visually supported answers from plausible but unsupported alternatives~\citep{mousi-etal-2026-correct,mousi2026said}.

Arabic provides an important setting for examining these issues because systems must handle MSA and regional dialects while grounding predictions in culturally specific content~\cite{al2025landscape}. Dallah~\citep{alwajih-etal-2024-dallah} highlights the importance of dialect-aware Arabic multimodal modeling, while CAMEL-Bench~\citep{ghaboura-etal-2025-camel} shows remaining gaps in Arabic multimodal performance. However, evaluation of culturally grounded VQA and hallucination across Arabic varieties remains limited, particularly for settings where questions are presented in spoken rather than written form.

The same concerns extend to text-to-image generation. Models may produce visually plausible images while omitting or distorting culturally important details. Standard measures of image quality and text-image alignment often fail to capture these errors or align with human judgments of cultural faithfulness~\citep{kannen2024beyond,bayramli-etal-2025-diffusion,nayak-etal-2025-culturalframes}. This is particularly relevant to Arabic and Gulf contexts, where distinct national characteristics may be reduced to generic regional representations~\citep{elsharif2024cultural,almarwani2025kingdomglimpses}.

To address these gaps, we introduce \textsc{\textbf{ImageEval 2026}}, a shared task on culturally grounded Arabic multimodal evaluation. It brings together two complementary tasks. \textsc{AynVQA} covers spoken visual question answering and hallucination detection, while \textsc{CRAI-Bench} evaluates cultural accuracy in Arabic text-to-image generation by assessing whether AI-generated images faithfully represent Arab cultural scenes. Overall, the tasks provide a unified Arabic-English evaluation of cultural grounding across multimodal understanding and generation.


\noindent\textbf{Findings:} The results highlight three main findings. 
\textbf{\textit{First}}, spoken VQA is more challenging in MSA than in English, with the best MSA accuracy 8.7 percentage points lower than the best English result. System analyses attribute much of this gap to Arabic ASR and the shift from synthetic training audio to human-recorded test speech. 
\textbf{\textit{Second}}, hallucination detection benefits from task-aware single-choice prediction. Systems that jointly select the visually grounded statement achieve substantially lower contrastive instability than independent true/false verification, with a zero-shot system ranking first in MSA. 
\textbf{\textit{Third}}, \textsc{CRAI-Bench} reveals that automated cultural evaluation can exploit non-visual structural cues. All submissions outperform the GPT-4o judge, yet a caption-version prior with limited visual evidence ranks first, while the hallucination-penalty dimension remains particularly difficult. 
Overall, these findings motivate stronger Arabic speech modeling, task-aligned hallucination evaluation, and cultural benchmarks that require direct visual grounding.
\section{Tasks and Datasets}
The shared task considers three complementary evaluation settings: grounding spoken questions in images, distinguishing visually supported content from culturally plausible hallucinations, and assessing cultural faithfulness in generated images. This section describes the corresponding tasks, datasets, and evaluation procedures.

\subsection{Task 1: AynVQA}
AynVQA is offered as two subtasks, each in English and MSA.
\begin{itemize}[noitemsep,topsep=0pt,leftmargin=*,labelsep=.5em]
\item \textbf{Task 1a (Spoken VQA):} \textit{Given an image and a spoken question with three spoken answer options, predict the index of the correct option.} Neither the question nor the options are provided as text.
\item \textbf{Task 1b (Hallucination Detection):} \textit{Given an image and three statements about it, judge each statement as true or false, where exactly one statement is grounded in the image and the other two are culturally plausible but visually unsupported.}
\end{itemize}

\begin{figure}[t]
\centering
\includegraphics[width=0.98\columnwidth]{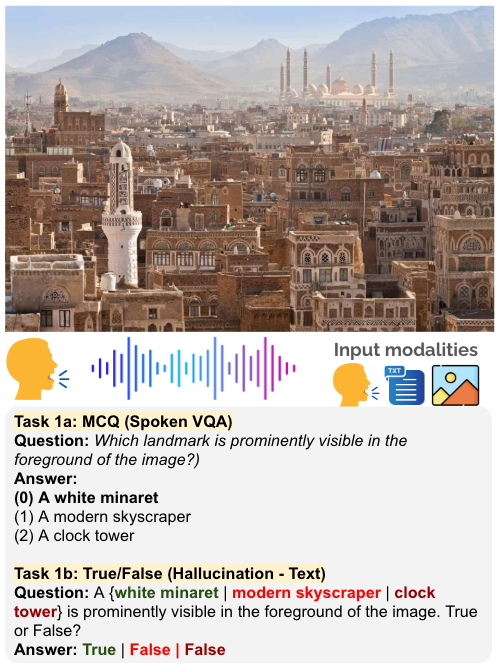}
\vspace{-0.2cm}
\caption{Example of Task~1: (1a) culturally grounded spoken visual QA with multiple-choice answers and (1b) image-grounded hallucination detection over true/false statements.}
\label{fig:ayn_sample}
\vspace{-0.3cm}
\end{figure}

\paragraph{Dataset.}
Task~1 is derived from OASIS~\citep{alam2025everydaymmqa}, which pairs images with spoken and textual QA instances in English and Arabic varieties across 18 MENA countries. Figure~\ref{fig:ayn_sample} illustrates the two subtasks.

\begin{table}[!tbh]
\centering
\small
\setlength{\tabcolsep}{2pt}
\scalebox{0.9}{
\begin{tabular}{lrlrl}
\toprule
\textbf{Split} & \textbf{\# Examples} & \textbf{Source} & \textbf{\# Countries} & \textbf{Speech} \\
\midrule
Train    & 3{,}000 & OASIS    & 18 & Voice-cloned \\
Dev      & 500     & OASIS    & 17 & Voice-cloned \\
Dev-test & 500     & OASIS    & 17 & Voice-cloned \\
Test     & 1{,}000 & M$^2$CQA & 13 & Human \\
\bottomrule
\end{tabular}
}
\vspace{-0.2cm}
\caption{Task~1 data splits by source, country coverage, and speech type.}
\label{tab:ayn-splits}
\vspace{-0.3cm}
\end{table}

The training, development, and dev-test splits are derived from OASIS, which contains multiple-choice (MCQ), open-ended (OEQ) and true/false questions. For Task~1a, we select a subset of the OASIS MCQ data for the shared task. For Task~1b, we adopt the approach proposed by \citet{mousi-etal-2026-correct}, converting each MCQ into three true/false statements using GPT-4.1: one true statement corresponding to the correct answer and two false statements derived from the distractors, as illustrated in Figure~\ref{fig:ayn_sample}. The test set is drawn from M$^2$CQA~\citep{mousi-etal-2026-correct}. The training, development, and dev-test splits use voice-cloned speech, whereas the test set uses human recordings, introducing variation in speakers and recording conditions. Each example is annotated with country, cultural category, and subcategory labels based on a taxonomy of nine categories and 31 subcategories, enabling fine-grained analysis across countries and cultural topics. To ensure cultural grounding and QA accuracy, a subset of the data was manually verified. Further details are provided in \cite{mousi-etal-2026-correct,alam2025everydaymmqa}. Table~\ref{tab:ayn-splits} summarizes the splits by size, source dataset, country coverage, and speech type.

\paragraph{Evaluation.}
For \textbf{Task~1a}, we evaluate systems using \textit{accuracy}, \textit{balanced accuracy}, and \textit{macro-F1}. For \textbf{Task~1b}, we adopt the contrastive evaluation proposed by \citet{mousi-etal-2026-correct}. Each example consists of a triplet with one visually grounded statement and two plausible but unsupported statements. The primary metric, \textbf{contrastive instability (CI)}, measures whether a system makes consistent predictions across the triplet rather than evaluating each statement independently. Let $N_{\text{partial}}$ denote the number of triplets with at least one correct prediction and $N_{\text{consistent}}$ the number of these triplets for which all three statements are correctly classified. We define 
$CI = 1 - \frac{N_{\text{consistent}}}{N_{\text{partial}}}.$
Lower CI indicates greater consistency: once a system correctly identifies part of a triplet, it should resolve the remaining statements correctly. We additionally report \textit{combined accuracy}, \textit{counterfactual hallucination rate} (CFHR), and separate accuracies for grounded and unsupported statements. Missing or unparseable predictions are counted as incorrect for both subtasks.

\paragraph{Baseline.}
We evaluate both subtasks in a zero-shot setting. For Task~1a, Qwen2.5-Omni-3B takes the image and spoken question as input and predicts the answer. For Task~1b, Qwen2.5-VL-3B takes the image and each statement as input and predicts whether the statement is grounded in the image.

\subsection{Task 2: CRAI-Bench}

\textsc{CRAI-Bench} evaluates automated methods for assessing cultural accuracy in text-to-image (T2I) generation. Each instance contains (i) a \textbf{reference image} depicting an authentic Qatari cultural scene, (ii) a \textbf{caption} used to prompt a T2I model, and (iii) the corresponding \textbf{generated image}. Given these inputs, systems predict scores along five dimensions of the \textbf{Cultural Representation Accuracy Index (CRAI)}, which is validated against human cultural judgments. 

\begin{figure*}[t]
\centering
\includegraphics[width=0.9\textwidth]{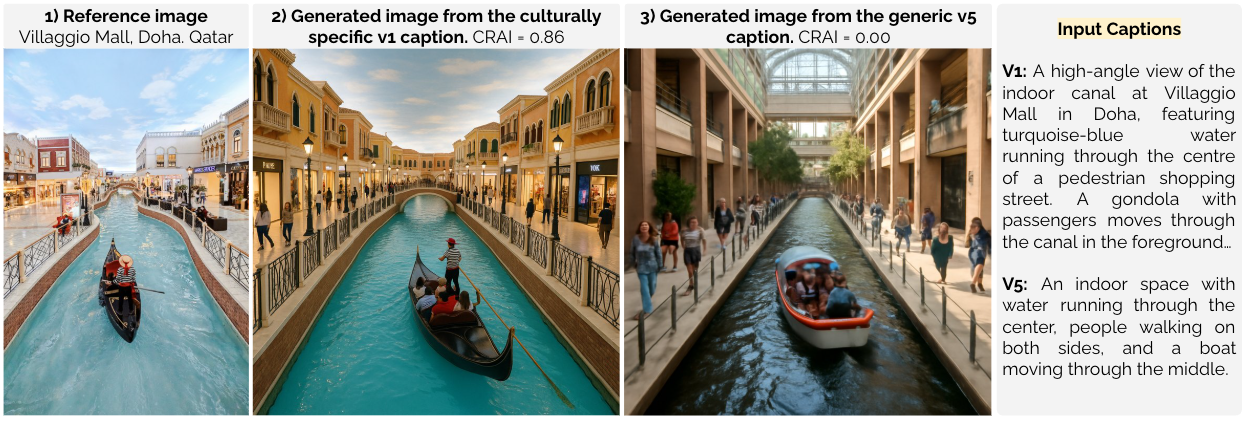}
\vspace{-0.2cm}
\caption{Example from \textsc{CRAI-Bench} (Task 2). Captions with decreasing cultural specificity generate substantially different representations of the same reference scene, reflected in their human-annotated CRAI scores.}
\label{fig:crai_sample}
\vspace{-0.3cm}
\end{figure*}

\begin{table}[!tbh]
\centering
\small
\setlength{\tabcolsep}{3pt}
\scalebox{0.82}{%
\begin{tabular}{p{1.65cm}ccp{4.25cm}}
\toprule
\textbf{Dimension} & \textbf{Code} & \textbf{Weight} & \textbf{Description} \\
\midrule
Cultural Element Accuracy & CEA & 0.30 & Whether expected cultural elements are present and correctly depicted \\
Contextual Coherence      & CC  & 0.20 & Whether elements appear in culturally appropriate settings \\
Cultural Specificity      & CS  & 0.20 & How specific the depiction is to the target culture \\
Cultural Integrity        & CI  & 0.20 & Whether the representation is truthful and free of distortion \\
Hallucination Penalty     & HP  & $-0.10$ & Culturally incorrect elements not supported by the caption \\
\bottomrule
\end{tabular}%
}
\vspace{-0.2cm}
\caption{CRAI dimensions, weights, and descriptions.}
\label{tab:crai-dimensions}
\vspace{-0.3cm}
\end{table}

\paragraph{Dataset.}
The \textsc{CRAI-Bench} dataset consists of 40 reference images depicting authentic Qatari cultural scenes across three categories: people and traditional attire, architecture and built environment, and objects. The images were collected from open-access platforms, including Pexels, Pixabay, and Unsplash, and their cultural authenticity was verified by native Qatari cultural consultants. Figure~\ref{fig:crai_sample} shows an example of a reference image and generated images.
Each reference image is paired with five English captions (v1 to v5) of decreasing cultural specificity. Each caption is then used to generate one image with \texttt{gpt-image-1} at a resolution of $1024\times1024$ pixels, yielding 200 caption-generated image instances in total. For each instance, human annotators evaluate the corresponding reference image, caption, and generated image using the CRAI rubric, assigning scores for the five dimensions and the composite CRAI score. The same instances are also evaluated by GPT-4o as an automated baseline.
The dataset is split by reference image into 60/20/20 train, development, and test partitions, stratified by cultural category. This ensures that captions and generated images associated with the same reference image do not appear across different splits. Table~\ref{tab:crai-splits} summarizes the split sizes.

\begin{table}[!tbh]
\centering
\small
\begin{tabular}{lcc}
\toprule
\textbf{Split} & \textbf{\# Ref. Images} & \textbf{\# Gen. Images} \\
\midrule
Train & 24 & 120 \\
Dev   & 8  & 40  \\
Test  & 8  & 40  \\
\midrule
Total & 40 & 200 \\
\bottomrule
\end{tabular}
\vspace{-0.2cm}
\caption{\textsc{CRAI-Bench} data splits by number of reference images and instances.}
\label{tab:crai-splits}
\vspace{-0.3cm}
\end{table}

\paragraph{Evaluation}

The primary evaluation metric is Spearman correlation~\citep{zar2005} ($\rho$) between predicted and human-annotated \texttt{CRAI\_composite} scores, with higher values indicating stronger agreement with human rankings. Mean Absolute Error (MAE) on the composite score is used as a secondary metric and tiebreaker, with lower values being better.

The official evaluation script computes both metrics and was distributed with the starter kit. During development, participants could evaluate their systems locally using the released development labels before submitting predictions to Codabench.

\paragraph{Baseline.}
We use GPT-4o as the baseline system. Given the reference image, caption, generated image, and CRAI rubric, the model predicts scores for the five CRAI dimensions. The composite CRAI score is then computed using the weighted formulation defined above.



\subsection{Competition Setup}
\label{sec:setup}

The shared task was conducted in two phases: \textbf{development} and \textbf{test}. During the development phase, participants developed their systems using the released training and development data and submitted predictions on the dev-test split to a live leaderboard. During the test phase, participants submitted predictions on the blind test set, which determined the official rankings. The test leaderboards remained hidden until the phase closed. Each team was allowed up to 16 test submissions, with a maximum of 10 per day, and the best submission was used for ranking.

Participants could use open- or closed-source models, external data, and pretrained models, provided that all resources were disclosed in the system description paper.

All Task~1 and Task~2 data are released under CC BY-NC-SA 4.0.\footnote{\href{https://huggingface.co/datasets/QCRI/ImageEval-ArabicNLP26}{Dataset}} The starter kits, evaluation scripts, and format checkers are also publicly 
available.\footnote{\href{https://github.com/ImageEval2026/ImageEval2026-tasks}{Task resources}}




\section{Results and Discussion}
\label{sec:results}


In Tables~\ref{tab:res-1a}, \ref{tab:res-1b}, and \ref{tab:res-2}, we report the official results on the blind test sets. Overall, the results highlight three patterns: spoken VQA remains more challenging in MSA than in English, task formulation plays an important role in hallucination detection, and cultural image evaluation can be influenced by structural cues beyond the visual content.

\begin{table}[t]
\centering
\footnotesize
\setlength{\tabcolsep}{3.5pt}
\scalebox{0.98}{
\begin{tabular}{@{}c l ccc@{}}
\toprule
\textbf{\#} & \textbf{Team} & \textbf{Acc.} & \textbf{B-Acc.} & \textbf{M-F1} \\
\midrule
\multicolumn{5}{@{}l}{\textit{\textbf{English}}} \\
\midrule
1 & Ahmed Ayman & 0.962 & 0.933 & 0.908 \\
2 & NYUAD & 0.961 & 0.919 & 0.872 \\
3 & CUET\_InferX & 0.912 & 0.877 & 0.805 \\
-- & \textit{Qwen2.5-Omni-3B (baseline)} & 0.594 & 0.619 & 0.483 \\
\midrule
\multicolumn{5}{@{}l}{\textit{\textbf{MSA}}} \\
\midrule
1 & NYUAD & 0.875 & 0.780 & 0.727 \\
2 & Ahmed Ayman & 0.834 & 0.721 & 0.686 \\
3 & Digilians & 0.656 & 0.575 & 0.504 \\
-- & \textit{Qwen2.5-Omni-3B (baseline)} & 0.191 & 0.339 & 0.169 \\
\bottomrule
\end{tabular}
}
\vspace{-0.2cm}
\caption{Task 1a (spoken visual QA) test results: accuracy, balanced
accuracy, and macro-F1. Qwen2.5-Omni-3B is the official baseline. For
reference, always answering the most frequent option position gives
0.539 accuracy in both languages.}
\label{tab:res-1a}
\vspace{-0.3cm}
\end{table}

\begin{table}[t]
\centering
\footnotesize
\setlength{\tabcolsep}{3.5pt}
\scalebox{0.98}{
\begin{tabular}{@{}c l ccc@{}}
\toprule
\textbf{\#} & \textbf{Team} & \textbf{CI}$\downarrow$ &
\textbf{Acc.}$_{Q^+}$ & \textbf{Acc.}$_{Q^-}$ \\
\midrule
\multicolumn{5}{@{}l}{\textit{\textbf{English}}} \\
\midrule
1 & Team Falcons & 0.029 & 0.971 & 0.986 \\
2 & Dynamos & 0.033 & 0.967 & 0.984 \\
3 & Team Tokenizers & 0.035 & 0.965 & 0.983 \\
4 & NYUAD & 0.041 & 0.959 & 0.980 \\
5 & alkhder & 0.051 & 0.949 & 0.975 \\
6 & Nile Nexus & 0.058 & 0.942 & 0.971 \\
7 & md\_faisal & 0.067 & 0.933 & 0.967 \\
8 & keslerady$^*$ & 0.113 & 0.887 & 0.944 \\
-- & \textit{Qwen2.5-VL-3B (baseline)} & 0.267 & 0.931 & 0.863 \\
\midrule
\multicolumn{5}{@{}l}{\textit{\textbf{MSA}}} \\
\midrule
1 & Ahmed Younis & 0.036 & 0.964 & 0.982 \\
2 & NYUAD & 0.047 & 0.953 & 0.977 \\
3 & Dynamos & 0.056 & 0.944 & 0.972 \\
4 & alkhder & 0.096 & 0.904 & 0.952 \\
5 & md\_faisal & 0.101 & 0.899 & 0.950 \\
6 & keslerady$^*$ & 0.142 & 0.858 & 0.929 \\
-- & \textit{Qwen2.5-VL-3B (baseline)} & 0.428 & 0.868 & 0.790 \\
\bottomrule
\end{tabular}
}
\vspace{-0.2cm}
\caption{Task 1b (hallucination detection) test results, ranked by
contrastive instability (CI, lower is better), with accuracy on the
grounded ($Q^+$) and hallucinated ($Q^-$) statements. Qwen2.5-VL-3B is
the official baseline. $^*$: no system description paper was submitted.}
\label{tab:res-1b}
\vspace{-0.2cm}
\end{table}

\begin{table}[t]
\centering
\footnotesize
\setlength{\tabcolsep}{3.5pt}
\scalebox{0.98}{
\begin{tabular}{@{}c l cc@{}}
\toprule
\textbf{\#} & \textbf{Team} & $\boldsymbol\rho\uparrow$ & \textbf{MAE}$\downarrow$ \\
\midrule
1 & md\_faisal & 0.826 & 0.147 \\
2 & surgan\_jandial$^*$ & 0.808 & 0.151 \\
3 & DATALabKU & 0.804 & 0.141 \\
4 & Ahmed Younis & 0.781 & 0.145 \\
5 & Ahmed Ayman & 0.673 & 0.203 \\
-- & \textit{GPT-4o (baseline)} & 0.519 & 0.236 \\
\bottomrule
\end{tabular}
}
\vspace{-0.2cm}
\caption{Task 2 (CRAI-Bench) test results, ranked by Spearman
correlation ($\rho$) of the predicted composite score against the
human gold scores. Mean absolute error is reported as a secondary
metric and does not affect the ranking. The GPT-4o judge is the
official baseline. $^*$: no system description paper was submitted
}
\label{tab:res-2}
\vspace{-0.3cm}
\end{table}


\paragraph{Task 1a (Spoken VQA).}
The English track was closely matched, with the top two systems differing by only 0.001 in accuracy. MSA was more challenging, with the best accuracy reaching 0.875 compared with 0.962 for English. Analysis of results across participating systems suggests that much of this gap comes from Arabic ASR errors and differences between the synthetic training and development speech and human-recorded test speech (Appendix~\ref{app:devtest}). Nevertheless, all submitted systems substantially outperform the Qwen2.5-Omni-3B baseline. In MSA, the baseline also falls below the majority-position reference (0.191 vs.\ 0.539), highlighting the difficulty of the spoken Arabic setting.


\paragraph{Task 1b (Hallucination Detection).}
For the hallucination detection subtask, performance was strong across submitted systems, with the top three English systems differing by only 0.006 CI. All ranked systems achieved a counterfactual hallucination rate of zero, making consistency across the three statements the main differentiator. Systems that jointly selected the single visually grounded statement generally outperformed the baseline, which evaluates each statement independently, reducing CI from 0.267 to 0.029 in English and from 0.428 to 0.036 in MSA. Notably, the top MSA system used zero-shot prompting rather than task-specific fine-tuning. These results suggest that aligning the prediction formulation with the contrastive structure of the task can be as important as model adaptation.

\paragraph{Task 2 (CRAI-Bench).}
For CRAI-Bench, all submitted systems outperform the GPT-4o baseline, with the top four achieving Spearman correlations between 0.781 and 0.826. The highest-ranked system uses a caption-version prior with CLIP-based tie-breaking, while DATALabKU~\cite{alkulaib-etal-2026-datalabku} achieves the lowest MAE. The strong performance of systems relying only partly on visual evidence reveals an important property of the benchmark, as human CRAI scores are strongly associated with caption specificity, making caption-version information highly predictive. This finding suggests that future versions should reduce such structural cues and place greater emphasis on direct visual assessment. The hallucination-penalty dimension also remains particularly challenging, with participant correlations near zero and a GPT-4o correlation of $-0.04$.

\section{System Descriptions}
\label{sec:systems}


Participating teams explored a range of approaches, including zero-shot prompting, task-specific fine-tuning, ASR-based pipelines, ensembling, and score calibration. In Table~\ref{tab:systems-overview}, we summarize the main models, adaptation strategies, and methods used across the three tasks. We briefly describe each submitted system below.
\begin{table*}[t]
\centering
\small
\setlength{\tabcolsep}{3.2pt}
\scalebox{0.85}{
\begin{tabular}{@{}l|cc|ccccccc|ccc|l@{}}
\toprule
\multirow{2}{*}{\textbf{Team}} &
\multicolumn{2}{c|}{\textbf{Track}} &
\multicolumn{7}{c|}{\textbf{Model / component}} &
\multicolumn{3}{c|}{\textbf{Adaptation}} &
\multicolumn{1}{c}{\textbf{Main method}} \\
&
\rot{\textbf{EN}} &
\rot{\textbf{MSA}} &
\rot{\textbf{Qwen-VL}} &
\rot{\textbf{Qwen-Omni}} &
\rot{\textbf{ASR}} &
\rot{\textbf{Gemini}} &
\rot{\textbf{GPT}} &
\rot{\textbf{Claude}} &
\rot{\textbf{CLIP}} &
\rot{\textbf{ZS}} &
\rot{\textbf{FT}} &
\rot{\textbf{Ens./Cal.}} &
\\
\midrule
\multicolumn{14}{@{}l}{\textit{\textbf{Task 1a: Spoken Visual Question Answering}}} \\
\midrule Ahmed Ayman & 1 & 2 & & & \cmark & & \cmark & & & \cmark & & & ASR--LLM cascade with ordinal answer labels\\
\rowcolor{gray!10} NYUAD & 2 & 1 & & & \cmark & \cmark & & & & \cmark & & & Comparison of native-audio and ASR-based pipelines\\
CUET InferX
& 3 &
& & \cmark & & & & &
& & \cmark &
& LoRA adaptation of the language backbone; audio and vision encoders frozen \\
\rowcolor{gray!10} Digilians & & 3 & \cmark & & \cmark & & & & & \cmark & & & ASR, constrained parsing, repair, and joint visual selection\\
\midrule
\multicolumn{14}{@{}l}{\textit{\textbf{Task 1b: Image-grounded Hallucination Detection}}} \\
\midrule Ahmed Younis & & 1 & & & & \cmark & & & & \cmark & & & Selection of the single visually grounded statement\\
\rowcolor{gray!10}
alkhder
& 5 & 4
& \cmark & & & & & &
& \cmark & & \cmark
& Rotation-averaged selection using next-token probabilities \\
Dynamos
& 2 & 3
& \cmark & & & & & &
& & \cmark & \cmark
& Constrained selection with cross-size model ensembling \\
\rowcolor{gray!10}
md\_faisal
& 7 & 5
& \cmark & & & & & &
& \cmark & &
& Forced-choice prediction from candidate-token scores \\
Nile Nexus
& 6 &
& \cmark & & & & & &
& \cmark & & \cmark
& Single-choice prediction with confidence-based paraphrase voting \\
\rowcolor{gray!10} NYUAD & 4 & 2 & & & & \cmark & & & & \cmark & & & Direct zero-shot statement verification\\
Team Falcons
& 1 &
& \cmark & & & \cmark & \cmark & &
& & \cmark & \cmark
& Fine-tuned classifier with consensus verification for uncertain cases \\
\rowcolor{gray!10}
Team Tokenizers
& 3 &
& \cmark & & & & & &
& & \cmark &
& Answer-first prediction with structured visual checks \\
\midrule
\multicolumn{14}{@{}l}{\textit{\textbf{Task 2: CRAI-Bench}}} \\
\midrule Ahmed Ayman & \multicolumn{2}{c|}{5} & & & & & \cmark & \cmark & & \cmark & & \cmark & Multiple judges with score calibration and tie-breaking\\
\rowcolor{gray!10} Ahmed Younis & \multicolumn{2}{c|}{4} & & & & \cmark & & & & \cmark & & \cmark & Caption-version prior combined with vision-based judges\\
DATALabKU
& \multicolumn{2}{c|}{3}
& & & & & \cmark & &
& \cmark & & \cmark
& Cultural evidence extraction followed by score regression \\
\rowcolor{gray!10}
md\_faisal
& \multicolumn{2}{c|}{1}
& & & & & & & \cmark
& & & \cmark
& Caption-version prior with CLIP-based tie-breaking \\
surgan\_jandial$^\dagger$ & \multicolumn{2}{c|}{2} & & & & & \cmark & & & \cmark & & \cmark & VLM judging with calibration and re-evaluation of near ties\\
\bottomrule
\end{tabular}
}
\vspace{-0.2cm}
\caption{Overview of submitted ImageEval 2026 systems. EN and MSA denote the English and Modern Standard Arabic tracks, respectively, for Tasks~1a and~1b. Task~2 (CRAI-Bench) consists of a single track and is therefore not divided by language. ZS: zero-shot prompting; FT: task-specific fine-tuning; Ens./Cal.: ensembling, voting, or score calibration. Model columns indicate components used anywhere in the submitted system. Numbers in the track columns give the team's position in the official ranking. $\dagger$System information obtained from the submission form; no system paper was submitted.}
\label{tab:systems-overview}
\vspace{-0.3cm}
\end{table*}

\paragraph{Dynamos~\cite{yamin-etal-2026-dynamos}}
The team participated in Task 1b for both English and Modern Standard Arabic (MSA). They reformulated hallucination detection as a constrained selection problem in which exactly one of three candidate statements is selected as visually grounded. Starting from zero-shot Qwen2.5-VL, they applied low-rank fine-tuning and trained multiple adapters with different random seeds. Their final system combined score averaging with an ensemble of 7B and 32B models, achieving CI scores of 0.033 for English and 0.056 for MSA.

\paragraph{NYUAD~\cite{AlDahoul-etal-2026-NYUAD}}
The team addressed both spoken visual question answering and image-grounded hallucination detection. Their framework jointly processes visual and language inputs and was evaluated in both English and MSA. The study compared different language model architectures and examined the additional challenges introduced by Arabic speech, text processing, and limited culturally grounded multimodal resources.

\paragraph{Nile Nexus~\cite{zaman-etal-2026-nile-nexus}}
Participating in Task 1b, the team used Qwen2.5-VL-7B-Instruct with 4-bit quantization and reformulated the task as selecting the single visually grounded statement. They also introduced confidence-based refinement through paraphrase voting. Their final system achieved an accuracy of 0.942 and a CI of 0.058 on the English track.

\paragraph{DATALabKU~\cite{alkulaib-etal-2026-datalabku}}
The team participated in Task 2 and proposed a reference-based framework for evaluating cultural accuracy in generated images. GPT-5.4 was used to extract cultural evidence from the reference image, caption, and generated image. These features were then mapped to CRAI scores using a lightweight regressor, improving Spearman correlation over direct GPT-5.4 scoring on the development set.

\paragraph{Team Falcons~\cite{ma-etal-2026-falcons}}
The team participated in the English track of Task 1b and formulated the task as three-way image-conditioned classification. Qwen3-VL-8B was fine-tuned using 4-bit QLoRA, while uncertain predictions were checked using Gemini 3.6 Flash and GPT-5.4-mini. A prediction was changed only when both verifier models agreed, improving the final CI to 0.029 and placing the system first on the English leaderboard.

\paragraph{Ahmed Ayman~\cite{mohamed-etal-2026-ahmedayman}}
The team participated in Task 1a and Task 2. For spoken VQA, they examined the effects of answer representation and speech recognition, with ordinal answer labels outperforming digit-based labels. Their system achieved accuracies of 0.962 for English and 0.834 for MSA. For CRAI-Bench, they used multiple judges with calibration and tie-breaking, achieving a Spearman correlation of 0.673.

\paragraph{md\_faisal~\cite{sheikh-2026-mdfaisal}}
The team participated in Task 1b for English and MSA, as well as Task 2, using zero-shot methods. For Task 1b, they replaced independent true/false predictions with a single three-way forced-choice formulation based on candidate-token logits. For CRAI-Bench, they used a caption-version prior with tie-breaking based on frozen CLIP features. The resulting system achieved a Spearman correlation of 0.8258 and ranked first on Task 2.

\paragraph{Team Tokenizers~\cite{hoque-etal-2026-team-tokenizers}}
Participating in the English track of Task 1b, the team proposed HALDETECT, which treats the three candidate statements as a single contrastive grounding decision. Their best system fine-tuned Qwen2.5-VL-7B-Instruct using 4-bit QLoRA while keeping the vision encoder frozen. The system achieved a CI of 0.035 and ranked third on the English leaderboard.

\paragraph{alkhder~\cite{pasa-etal-2026-alkhder}}
The team participated in Task 1b for both English and MSA using a pre-trained 7B vision--language model in a zero-shot setting. They selected the single visually supported statement using next-token log-probabilities and reduced positional bias through a Latin-square rotation over candidate orderings. The system achieved CI scores of 0.051 for English and 0.096 for MSA.

\paragraph{CUET InferX~\cite{semon-etal-2026-cuet-inferx}}
The team participated in the English track of Task 1a. They fine-tuned Qwen2.5-Omni-3B using LoRA on the language reasoning backbone while keeping the audio and vision encoders frozen. The resulting end-to-end audio--visual system achieved 91.20\% test accuracy and ranked third in the English track.

\paragraph{Ahmed Younis~\cite{younis-etal-2026-ahmedyounis}}
The team participated in the MSA track of Task 1b and in Task 2. For CRAI-Bench, they combined a frequency-based prior with two vision-based judges and optimized the blending weights against Spearman correlation, achieving 0.7806 on the test set. For Task 1b, they reformulated the task as selecting the single grounded statement, achieving a combined accuracy of 0.964 and ranking first on the MSA track.

\paragraph{Digilians~\cite{hassan-etal-2026-digilians}}
The team participated in the MSA track of Task 1a using a modular spoken VQA pipeline. Their approach first performed speech recognition, then extracted the question and answer choices before applying visual multiple-choice reasoning. A larger speech recognition model was used only when the initial transcript was unreliable. The system achieved 79.0\% accuracy on the development set and 65.6\% on the blind test set.

\section{Related Work}
\label{sec:related_work}

\paragraph{Spoken Visual Question Answering.}

Prior work on spoken QA has examined the effects of ASR errors, conversational context, and multilingual speech~\citep{lee2018spoken,you2020towards,alam2025spokennativqa,ali2026wasilinthewildarabicspoken,wang2026mmsu}. These benchmarks, however, do not require visual grounding. More recent datasets such as \textsc{TM-PathVQA} and \textsc{TM-VQA} combine spoken questions with images across multiple languages~\citep{rajkhowa24_interspeech,chowdhury-etal-2025-towards}, but rely mainly on synthesized speech and do not 
target culturally grounded interaction.

Arabic multimodal QA has so far focused largely on image-text settings. \textsc{VAQA}, \textsc{CAMEL-Bench}, \textsc{PEARL}, \textsc{JEEM}, and \textsc{AraVQA} cover domains, cultural knowledge, dialects, and factoid questions~\citep{kamel2023vaqa,ghaboura-etal-2025-camel,alwajih2025pearl,kadaoui-etal-2026-jeem,alrowili-etal-2026-aravqa}. \textsc{OASIS} extends this line of work to speech with 3.7M spoken questions across 18 Arab countries, covering MSA and dialectal Arabic~\citep{alam2025everydaymmqa}. 

\paragraph{Hallucination Benchmarks and Metrics.}

Vision-language hallucination is commonly divided into \emph{faithfulness hallucination}, where outputs conflict with the visual input, and \emph{factuality hallucination}, where they contradict external knowledge~\citep{chen2026survey}. ImageEval focuses on faithfulness hallucination. Existing benchmarks such as POPE, ROPE, FGHE, RAH-Bench, R-Bench, and AutoHallusion evaluate unsupported objects, attributes, relations, and spatial setups~\citep{rope,wang2023mitigating,wu-etal-2024-autohallusion}. Broader suites include LongHalQA, AMBER, and MERLIM~\citep{qiu2024longhalqa,wang2024amber,Villa_2025_CVPR}, while CHAIR, OpenCHAIR, CC-EVAL, and NOPE evaluate hallucination in open-ended generation~\citep{rohrbach-etal-2018-object,ben-kish-etal-2024-mitigating,zhai2024halle,lovenia-etal-2024-negative}.


Most of these benchmarks use accuracy, F1, or object-level precision and recall. Such measures do not clearly distinguish visual-recognition failures from cases where a model recognizes the image correctly but accepts a plausible unsupported alternative. Many also rely on English-language or Western-centric resources such as MS COCO~\citep{mscoco}. Although CVQA introduces culturally diverse visual questions~\citep{romero2024cvqa}, hallucination evaluation for Arabic 
and the MENA region 
remains limited~\cite{alansari2025arahallueval,mousi-etal-2026-correct}.

\paragraph{Cultural Evaluation of Text-to-Image Generation.}

Recent text-to-image benchmarks increasingly evaluate cultural accuracy in addition to visual quality and prompt alignment. CUBE, CULTDIFF, CulturalFrames, and RusCode examine culturally salient content across countries and settings~\citep{kannen2024beyond,bayramli-etal-2025-diffusion,nayak-etal-2025-culturalframes, vasilev-etal-2025-ruscode}. CULTIVate and MOSAIG extend this direction to social activities and multicultural scenes~\citep{malakouti2026culture,bhalerao-etal-2026-cultures}. Related efforts include the Cultural Relevance Index for Arabic culture~\citep{elsharif2024cultural}, its use in Ara-Pic~\citep{elsharif2025arapic}, CAIRe's knowledge-grounded cultural judgments~\citep{yayavaram-etal-2026-caire}, and human-centered evaluation rubrics~\citep{johnson2026evaluating}.

\noindent\textsc{\textbf{ImageEval 2026}} builds on these efforts by bringing culturally grounded understanding and culturally faithful generation into a unified Arabic- and MENA-centered evaluation setting. It jointly evaluates culturally grounded spoken visual QA, image-grounded hallucination detection, and the cultural fidelity of generated images in bilingual Arabic-English settings.



\section{Conclusion}
\label{sec:conclusion}

We presented \textsc{ImageEval 2026}, a shared task on culturally grounded Arabic multimodal evaluation. The shared task combines \textsc{AynVQA}, which evaluates spoken visual question answering and image-grounded hallucination detection in English and MSA, with \textsc{CRAI-Bench}, which evaluates cultural accuracy in generated images. A total of 14 teams participated in the test phase, with 12 teams submitting system description papers.
The results highlight three main findings. First, spoken VQA remains more challenging in MSA than in English, with Arabic ASR contributing substantially to the performance gap. Second, hallucination detection benefits from task-aware single-choice formulations that directly identify the visually grounded statement. Third, \textsc{CRAI-Bench} shows that caption-specificity cues can strongly influence cultural evaluation, allowing systems with limited use of visual evidence to perform well. These findings motivate further work on robust Arabic speech processing, visually grounded hallucination detection, and cultural evaluation methods that rely more directly on image content. We release the datasets and evaluation resources to support future research on culturally grounded Arabic multimodal systems. 
For future work, we plan to expand the benchmark with broader Arabic dialect coverage, more diverse speech and image-generation conditions, and stronger controls against structural cues. 



\section*{Limitations}
For Task~1a, the training, development, and dev-test audio is synthesized, while the test set uses human recordings, introducing a challenge in speech conditions that particularly affects MSA performance. Task~1b uses a constrained setting with exactly one grounded statement and two unsupported alternatives, so the results may not generalize to less structured hallucination scenarios. Task~2 is limited to Qatari cultural content and contains 40 reference images and 200 generated images, all produced using a single text-to-image model. Its five caption versions also introduce a strong association between caption specificity and human scores, which allowed systems using caption-version information to perform well with limited use of visual evidence. Future editions should broaden the cultural and geographic coverage, include more varied speech and generation sources, and reduce structural cues that can be exploited without fully evaluating the image.

\section*{Societal/Broader Impact}
ImageEval 2026 aims to improve the evaluation of multimodal systems for Arab cultural contexts across speech, text, and image generation. By releasing shared data, evaluation scripts, and baselines, the benchmark can lower the barrier to developing and comparing Arabic multimodal systems and help identify weaknesses that may otherwise remain hidden in predominantly English-centric evaluation. Our results highlight notable gaps, particularly in Arabic speech recognition and the cultural accuracy of generated images, which may affect the reliability and inclusiveness of systems deployed for Arabic-speaking users. At the same time, benchmark scores should not be interpreted as a complete measure of cultural competence, as Arab societies are linguistically, geographically, and culturally diverse. We therefore encourage future work to expand coverage across countries, dialects, communities, and cultural perspectives, and to use the benchmark as one component of broader human-centered evaluation.



\bibliography{bibliography/bibliography,bibliography/references}

@inproceedings{alansari2025arahallueval,
  title={{AraHalluEval}: A fine-grained hallucination evaluation framework for Arabic LLMs},
  author={Alansari, Aisha and Luqman, Hamzah},
  booktitle={Proceedings of The Third Arabic Natural Language Processing Conference},
  pages={148--161},
  year={2025}
}

@inproceedings{
wang2026mmsu,
title={{MMSU}: A Massive Multi-task Spoken Language Understanding and Reasoning Benchmark},
author={Dingdong WANG and Junan Li and Jincenzi Wu and Dongchao Yang and Xueyuan Chen and Tianhua Zhang and Helen M. Meng},
booktitle={The Fourteenth International Conference on Learning Representations},
year={2026},
url={https://openreview.net/forum?id=yHzCDP1tXw}
}

@inproceedings{ali2026wasilinthewildarabicspoken,
  title = {{WASIL}: In-the-Wild Arabic Spoken Interactions with LLMs},
  author = {Ali, Zien Sheikh and Mubarak, Hamdy and Jung, Soon-Gyo and Bhatti, Hunzalah Hassan and Alam, Firoj and Chowdhury, Shammur Absar},
  booktitle = {Proceedings of Interspeech 2026},
  year = {2026},
  address = {Sydney, Australia},
}

@inproceedings{romero2024cvqa,
author = {Romero, David and Lyu, Chenyang and Wibowo, Haryo Akbarianto and Lynn, Teresa and Hamed, Injy and Kishore, Aditya Nanda and Mandal, Aishik and Dragonetti, Alina and Abzaliev, Artem and Tonja, Atnafu Lambebo and Balcha, Bontu Fufa and Whitehouse, Chenxi and Salamea, Christian and Velasco, Dan John and Adelani, David Ifeoluwa and Le Meur, David and Villa-Cueva, Emilio and Koto, Fajri and Farooqui, Fauzan and Belcavello, Frederico and Batnasan, Ganzorig and Vallejo, Gisela and Caulfield, Grainne and Ivetta, Guido and Song, Haiyue and Ademtew, Henok Biadglign and Maina, Hern{\'a}n and Lovenia, Holy and Azime, Israel Abebe and Cruz, Jan Christian Blaise and Gala, Jay and Geng, Jiahui and Ortiz-Barajas, Jesus-German and Baek, Jinheon and Dunstan, Jocelyn and Alemany, Laura Alonso and Nagasinghe, Kumaranage Ravindu Yasas and Benotti, Luciana and D'Haro, Luis Fernando and Viridiano, Marcelo and Estecha-Garitagoitia, Marcos and Cabrera, Maria Camila Buitrago and Rodr{\'i}guez-Cantelar, Mario and Jouitteau, M{\'e}lanie and Mihaylov, Mihail and Etori, Naome and Imam, Mohamed Fazli Mohamed and Adilazuarda, Muhammad Farid and Gochoo, Munkhjargal and Otgonbold, Munkh-Erdene and Niyomugisha, Olivier and Silva, Paula M{\'o}nica and Chitale, Pranjal and Dabre, Raj and Chevi, Rendi and Zhang, Ruochen and Diandaru, Ryandito and Cahyawijaya, Samuel and G{\'o}ngora, Santiago and Jeong, Soyeong and Purkayastha, Sukannya and Kuribayashi, Tatsuki and Clifford, Teresa and Jayakumar, Thanmay and Torrent, Tiago Timponi and Ehsan, Toqeer and Araujo, Vladimir and Kementchedjhieva, Yova and Burzo, Zara and Lim, Zheng Wei and Yong, Zheng Xin and Ignat, Oana and Nwatu, Joan and Mihalcea, Rada and Solorio, Thamar and Aji, Alham Fikri},
title = {CVQA: culturally-diverse multilingual visual question answering benchmark},
year = {2024},
isbn = {9798331314385},
publisher = {Curran Associates Inc.},
address = {Red Hook, NY, USA},
booktitle = {Proceedings of the 38th International Conference on Neural Information Processing Systems},
articleno = {366},
numpages = {27},
location = {Vancouver, BC, Canada},
series = {NIPS '24}
}

@inproceedings{mousi-etal-2026-correct,
  title = {Once Correct, Still Wrong: Counterfactual Hallucination in Multilingual Vision-Language Models},
  author = {Mousi, Basel and Dalvi, Fahim and Chowdhury, Shammur Absar and Alam, Firoj and Durrani, Nadir},
  editor = {Liakata, Maria and Moreira, Viviane P. and Zhang, Jiajun and Jurgens, David},
  booktitle = {Findings of the {A}ssociation for {C}omputational {L}inguistics: {ACL} 2026},
  month = jul,
  year = {2026},
  address = {San Diego, California, United States},
  publisher = {Association for Computational Linguistics},
  url = {https://aclanthology.org/2026.findings-acl.234/},
  pages = {4763--4788},
  isbn = {979-8-89176-395-1},
}

@inproceedings{mousi2026said,
  title = {Said Aloud, Read Different: Cross-Modal Instability in Multimodal Models},
  author = {Mousi, Basel and Dalvi, Fahim and Chowdhury, Shammur and Alam, Firoj and Durrani, Nadir},
  booktitle = {Proceedings of Interspeech 2026},
  year = {2026},
  address = {Sydney, Australia},
  note = {accepted},
}

@inproceedings{yayavaram-etal-2026-caire,
    title = "{CAIRE}: Cultural Attribution of Images with Retrieval",
    author = "Yayavaram, Arnav  and
      Yayavaram, Siddharth  and
      Khanuja, Simran  and
      Saxon, Michael  and
      Neubig, Graham",
    editor = "Demberg, Vera  and
      Inui, Kentaro  and
      Marquez, Llu{\'i}s",
    booktitle = "Proceedings of the 19th Conference of the {E}uropean Chapter of the {A}ssociation for {C}omputational {L}inguistics (Volume 1: Long Papers)",
    month = mar,
    year = "2026",
    address = "Rabat, Morocco",
    publisher = "Association for Computational Linguistics",
    url = "https://aclanthology.org/2026.eacl-long.389/",
    doi = "10.18653/v1/2026.eacl-long.389",
    pages = "8320--8338",
    ISBN = "979-8-89176-380-7"
}

@inproceedings{johnson2026evaluating,
  title={Evaluating AI-Generated Images of Cultural Artifacts with Community-Informed Rubrics},
  author={Johnson, Nari and Sudharsan, Deepthi and Hamna and Dalal, Samantha and Holroyd, Theo and Thieme, Anja and Heidari, Hoda and Massiceti, Daniela and Wortman Vaughan, Jennifer and Morrison, Cecily},
  booktitle={The 2026 ACM Conference on Fairness, Accountability, and Transparency},
  pages={714--774},
  year={2026}
}

@inproceedings{alrowili-etal-2026-aravqa,
    title = "{A}ra{VQA}: Building a New {A}rabic Factoid Visual Question Answering Dataset from {W}ikipedia",
    author = "Alrowili, Sultan  and
      Samih, Younes  and
      Freihat, Abed Alhakim  and
      Eswaran, Mathan Kumar",
    editor = "Liakata, Maria  and
      Moreira, Viviane P.  and
      Zhang, Jiajun  and
      Jurgens, David",
    booktitle = "Proceedings of the 64th Annual Meeting of the {A}ssociation for {C}omputational {L}inguistics (Volume 1: Long Papers)",
    month = jul,
    year = "2026",
    address = "San Diego, California, United States",
    publisher = "Association for Computational Linguistics",
    url = "https://aclanthology.org/2026.acl-long.91/",
    doi = "10.18653/v1/2026.acl-long.91",
    pages = "2026--2042",
    ISBN = "979-8-89176-390-6"
}

@inproceedings{kadaoui-etal-2026-jeem,
    title = "{JEEM}: Vision-Language Understanding in Four {A}rabic Dialects",
    author = "Kadaoui, Karima  and
      Atwany, Hanin  and
      Al-Ali, Hamdan  and
      Mohamed, Abdelrahman  and
      Mekky, Ali  and
      Tilga, Sergei  and
      Fedorova, Natalia  and
      Artemova, Ekaterina  and
      Aldarmaki, Hanan  and
      Kementchedjhieva, Yova",
    editor = "Demberg, Vera  and
      Inui, Kentaro  and
      Marquez, Llu{\'i}s",
    booktitle = "Findings of the {A}ssociation for {C}omputational {L}inguistics: {EACL} 2026",
    month = mar,
    year = "2026",
    address = "Rabat, Morocco",
    publisher = "Association for Computational Linguistics",
    url = "https://aclanthology.org/2026.findings-eacl.18/",
    doi = "10.18653/v1/2026.findings-eacl.18",
    pages = "331--354",
    ISBN = "979-8-89176-386-9"
}

@article{kamel2023vaqa,
  author  = {Kamel, Sarah M. and Hassan, Shimaa I. and Elrefaei, Lamiaa},
  title   = {{VAQA}: Visual Arabic Question Answering},
  journal = {Arabian Journal for Science and Engineering},
  year    = {2023},
  volume  = {48},
  pages   = {10803--10823},
  doi     = {10.1007/s13369-023-07687-y},
  url     = {https://doi.org/10.1007/s13369-023-07687-y}
}

@inproceedings{chowdhury-etal-2025-towards,
    title = "Towards Multilingual spoken Visual Question Answering system using Cross-Attention",
    author = "Chowdhury, Amartya Roy  and
      Rajkhowa, Tonmoy  and
      Sharma, Sanjeev",
    editor = "Rambow, Owen  and
      Wanner, Leo  and
      Apidianaki, Marianna  and
      Al-Khalifa, Hend  and
      Eugenio, Barbara Di  and
      Schockaert, Steven",
    booktitle = "Proceedings of the 31st International Conference on Computational Linguistics",
    month = jan,
    year = "2025",
    address = "Abu Dhabi, UAE",
    publisher = "Association for Computational Linguistics",
    url = "https://aclanthology.org/2025.coling-main.615/",
    pages = "9165--9175"
}

@inproceedings{rajkhowa24_interspeech,
  title     = {{TM-PATHVQA: 90000+ Textless Multilingual Questions for Medical Visual Question Answering}},
  author    = {Tonmoy Rajkhowa and Amartya Roy Chowdhury and Sankalp Nagaonkar and Achyut Mani Tripathi and Mahadeva Prasanna},
  year      = {2024},
  booktitle = {{Interspeech 2024}},
  pages     = {4034--4038},
  doi       = {10.21437/Interspeech.2024-1036},
  issn      = {2958-1796},
}

@inproceedings{kannen2024beyond,
  title     = {Beyond Aesthetics: Cultural Competence in Text-to-Image Models},
  author    = {Kannen, Nithish and Ahmad, Arif and Andreetto, Marco and
               Prabhakaran, Vinodkumar and Prabhu, Utsav and
               Dieng, Adji Bousso and Bhattacharyya, Pushpak and Dave, Shachi},
  booktitle = {Advances in Neural Information Processing Systems},
  volume    = {37},
  pages     = {13716--13747},
  year      = {2024},
  doi       = {10.52202/079017-0439},
  url       = {https://proceedings.neurips.cc/paper_files/paper/2024/hash/18c669b80d1a8f589713b768bc8fe9a4-Abstract-Datasets_and_Benchmarks_Track.html}
}

@inproceedings{elsharif2024cultural,
  title     = {Cultural Relevance Index: Measuring Cultural Relevance
               in AI-Generated Images},
  author    = {Elsharif, Wala and Agus, Marco and Alzubaidi, Mahmoud and
               She, James},
  booktitle = {2024 IEEE 7th International Conference on Multimedia
               Information Processing and Retrieval (MIPR)},
  pages     = {410--416},
  year      = {2024},
  month     = oct,
  publisher = {IEEE},
  isbn      = {979-8-3503-5143-9},
  doi       = {10.1109/MIPR62202.2024.00071},
  url       = {https://ieeexplore.ieee.org/document/10707879/}
}

@inproceedings{bayramli-etal-2025-diffusion,
    title = "Diffusion Models Through a Global Lens: Are They Culturally Inclusive?",
    author = "Bayramli, Zahra  and
      Suleymanzade, Ayhan  and
      An, Na Min  and
      Ahmad, Huzama  and
      Kim, Eunsu  and
      Park, Junyeong  and
      Thorne, James  and
      Oh, Alice",
    editor = "Che, Wanxiang  and
      Nabende, Joyce  and
      Shutova, Ekaterina  and
      Pilehvar, Mohammad Taher",
    booktitle = "Proceedings of the 63rd Annual Meeting of the Association for Computational Linguistics (Volume 1: Long Papers)",
    month = jul,
    year = "2025",
    address = "Vienna, Austria",
    publisher = "Association for Computational Linguistics",
    url = "https://aclanthology.org/2025.acl-long.1503/",
    doi = "10.18653/v1/2025.acl-long.1503",
    pages = "31137--31155",
    ISBN = "979-8-89176-251-0"
}

@inproceedings{nayak-etal-2025-culturalframes,
  title     = {{CulturalFrames}: Assessing Cultural Expectation Alignment
               in Text-to-Image Models and Evaluation Metrics},
  author    = {Nayak, Shravan and Bhatia, Mehar and Zhang, Xiaofeng and
               Rieser, Verena and Hendricks, Lisa Anne and
               van Steenkiste, Sjoerd and Goyal, Yash and
               Stanczak, Karolina and Agrawal, Aishwarya},
  editor    = {Christodoulopoulos, Christos and Chakraborty, Tanmoy and
               Rose, Carolyn and Peng, Violet},
  booktitle = {Findings of the Association for Computational Linguistics:
               EMNLP 2025},
  month     = nov,
  year      = {2025},
  address   = {Suzhou, China},
  publisher = {Association for Computational Linguistics},
  pages     = {20918--20953},
  isbn      = {979-8-89176-335-7},
  doi       = {10.18653/v1/2025.findings-emnlp.1141},
  url       = {https://aclanthology.org/2025.findings-emnlp.1141/}
}

@inproceedings{vasilev-etal-2025-ruscode,
    title = "{R}us{C}ode: {R}ussian Cultural Code Benchmark for Text-to-Image Generation",
    author = "Vasilev, Viacheslav  and
      Agafonova, Julia  and
      Gerasimenko, Nikolai  and
      Kapitanov, Alexander  and
      Mikhailova, Polina  and
      Mironova, Evelina  and
      Dimitrov, Denis",
    editor = "Chiruzzo, Luis  and
      Ritter, Alan  and
      Wang, Lu",
    booktitle = "Findings of the Association for Computational Linguistics: NAACL 2025",
    month = apr,
    year = "2025",
    address = "Albuquerque, New Mexico",
    publisher = "Association for Computational Linguistics",
    url = "https://aclanthology.org/2025.findings-naacl.425/",
    doi = "10.18653/v1/2025.findings-naacl.425",
    pages = "7656--7672",
    ISBN = "979-8-89176-195-7"
}

@inproceedings{elsharif2025arapic,
  title     = {{Ara-Pic}: A Framework for Enhancing Arabic Cultural
               Representation in AI-Generated Images},
  author    = {Elsharif, Wala and Alzubaidi, Mahmoud and She, James and
               Agus, Marco},
  booktitle = {2025 IEEE International Conference on Multimedia and
               Expo Workshops (ICMEW)},
  pages     = {1--6},
  year      = {2025},
  publisher = {IEEE},
  doi       = {10.1109/ICMEW68306.2025.11152135},
  url       = {https://ieeexplore.ieee.org/document/11152135/}
}

@article{almarwani2025kingdomglimpses,
  title   = {{KingdomGlimpses}: Evaluating Saudi Cultural Representation
             Through Text-to-Image Models},
  author  = {Almarwani, Nada and Aloufi, Samah and Alkhereyf, Sakhar B. and
             Alhassoun, Manal and Almutery, Manal and Alshalawi, Nouf and
             Al-Thubaity, Abdulmohsen},
  journal = {IEEE Access},
  volume  = {13},
  pages   = {177822--177845},
  year    = {2025},
  doi     = {10.1109/ACCESS.2025.3619432},
  url     = {https://ieeexplore.ieee.org/document/11196757/}
}

@inproceedings{malakouti2026culture,
  title     = {Culture in Action: Evaluating Text-to-Image Models
               through Social Activities},
  author    = {Malakouti, Sina and Gong, Boqing and Kovashka, Adriana},
  booktitle = {The Fourteenth International Conference on
               Learning Representations},
  year      = {2026},
  url       = {https://openreview.net/forum?id=opG4m2U0Oo}
}

@inproceedings{bhalerao-etal-2026-cultures,
  title     = {When Cultures Meet: Multicultural Text-to-Image Generation},
  author    = {Bhalerao, Parth and Yalamarty, Mounika and Trinh, Brian and
               Ignat, Oana},
  editor    = {Liakata, Maria and Moreira, Viviane P. and Zhang, Jiajun and
               Jurgens, David},
  booktitle = {Findings of the Association for Computational Linguistics:
               ACL 2026},
  month     = jul,
  year      = {2026},
  address   = {San Diego, California, United States},
  publisher = {Association for Computational Linguistics},
  pages     = {35808--35828},
  isbn      = {979-8-89176-395-1},
  doi       = {10.18653/v1/2026.findings-acl.1783},
  url       = {https://aclanthology.org/2026.findings-acl.1783/}
}

@inproceedings{alwajih2025pearl,
  address = {Suzhou, China},
  author = {Alwajih, Fakhraddin  and
Magdy, Samar M.  and
El Mekki, Abdellah  and
Nacar, Omer  and
Nafea, Youssef  and
Abdelfadil, Safaa Taher  and
Yahya, Abdulfattah Mohammed  and
Luqman, Hamzah  and
Almarwani, Nada  and
Aloufi, Samah  and
Qawasmeh, Baraah  and
Atou, Houdaifa  and
Sibaee, Serry  and
Alsayadi, Hamzah A.  and
Al-Dhabyani, Walid  and
Al-shaibani, Maged S.  and
El aatar, Aya  and
Qandos, Nour  and
Alhamouri, Rahaf  and
Ahmad, Samar  and
AL-Ghrawi, Mohammed Anwar  and
Yacoub, Aminetou  and
AbuHweidi, Ruwa  and
Lemin, Vatimetou Mohamed  and
Abdel-Salam, Reem  and
Bashiti, Ahlam  and
Ammar, Adel  and
Alansari, Aisha  and
Ashraf, Ahmed  and
Alturayeif, Nora  and
Alcoba Inciarte, Alcides  and
Elmadany, AbdelRahim A.  and
Tourad, Mohamedou Cheikh  and
Berrada, Ismail  and
Jarrar, Mustafa  and
Shehata, Shady  and
Abdul-Mageed, Muhammad},
  booktitle = {Findings of the Association for Computational Linguistics: EMNLP 2025},
  doi = {10.18653/v1/2025.findings-emnlp.1254},
  editor = {Christodoulopoulos, Christos  and
Chakraborty, Tanmoy  and
Rose, Carolyn  and
Peng, Violet},
  isbn = {979-8-89176-335-7},
  pages = {23048--23079},
  publisher = {Association for Computational Linguistics},
  title = {Pearl: A Multimodal Culturally-Aware {A}rabic Instruction Dataset},
  url = {https://aclanthology.org/2025.findings-emnlp.1254/},
  year = {2025}
}

@article{alam2025everydaymmqa,
  title = {{OASIS}: A Multilingual and Multimodal Dataset for Culturally Grounded Spoken Visual QA},
  author = {Alam, Firoj and Shahroor, Ali Ezzat and Hasan, Md. Arid and Ali, Zien Sheikh and Bhatti, Hunzalah Hassan and Kmainasi, Mohamed Bayan and Chowdhury, Shammur Absar and Mousi, Basel and Dalvi, Fahim and Durrani, Nadir and Milic-Frayling, Natasa},
  journal = {arXiv preprint arXiv:2510.06371},
  year = {2025},
}

@article{al2025landscape,
  title = {The Landscape of Arabic Large Language Models},
  author = {Al-Khalifa, Shahad and Durrani, Nadir and Al-Khalifa, Hend and Alam, Firoj},
  journal = {Communications of the ACM},
  volume = {68},
  number = {10},
  pages = {54--61},
  year = {2025},
  publisher = {ACM New York, NY, USA},
}

@inproceedings{lee2018spoken,
  title     = {{Spoken SQuAD: A Study of Mitigating the Impact of Speech Recognition Errors on Listening Comprehension}},
  author    = {Chia-Hsuan Lee and Szu-Lin Wu and Chi-Liang Liu and Hung-yi Lee},
  year      = {2018},
  booktitle = {{Interspeech 2018}},
  pages     = {3459--3463},
  doi       = {10.21437/Interspeech.2018-1714},
  issn      = {2958-1796},
}

@inproceedings{you2020towards,
    title = "End-to-end Spoken Conversational Question Answering: Task, Dataset and Model",
    author = "You, Chenyu  and
      Chen, Nuo  and
      Liu, Fenglin  and
      Ge, Shen  and
      Wu, Xian  and
      Zou, Yuexian",
    editor = "Carpuat, Marine  and
      de Marneffe, Marie-Catherine  and
      Meza Ruiz, Ivan Vladimir",
    booktitle = "Findings of the Association for Computational Linguistics: NAACL 2022",
    month = jul,
    year = "2022",
    address = "Seattle, United States",
    publisher = "Association for Computational Linguistics",
    url = "https://aclanthology.org/2022.findings-naacl.91/",
    doi = "10.18653/v1/2022.findings-naacl.91",
    pages = "1219--1232"
}

@inproceedings{alam2025spokennativqa,
  title = {{SpokenNativQA}: Multilingual Everyday Spoken Queries for LLMs},
  author = {Alam, Firoj and Hasan, Md Arid and Chowdhury, Shammur Absar},
  booktitle = {Proceedings of the 26th Interspeech Conference (Interspeech 2025)},
  year = {2025},
  address = {Rotterdam, The Netherlands},
  month = aug,
  organization = {ISCA},
}

@inproceedings{rope,
 author = {Xuweiyi Chen and
Ziqiao Ma and
Xuejun Zhang and
Sihan Xu and
Shengyi Qian and
Jianing Yang and
David Fouhey and
Joyce Chai},
 bibsource = {dblp computer science bibliography, https://dblp.org},
 booktitle = {Advances in Neural Information Processing Systems 38: Annual Conference
on Neural Information Processing Systems 2024, NeurIPS 2024, Vancouver,
BC, Canada, December 10 - 15, 2024},
 editor = {Amir Globersons and
Lester Mackey and
Danielle Belgrave and
Angela Fan and
Ulrich Paquet and
Jakub M. Tomczak and
Cheng Zhang},
 title = {Multi-Object Hallucination in Vision Language Models},
 url = {http://papers.nips.cc/paper\_files/paper/2024/hash/4ea4a1ea4d9ff273688c8e92bd087112-Abstract-Conference.html},
 year = {2024}
}

@inproceedings{wang2023mitigating,
author = {Wang, Lei and He, Jiabang and Li, Shenshen and Liu, Ning and Lim, Ee-Peng},
title = {Mitigating Fine-Grained Hallucination by Fine-Tuning Large Vision-Language Models with Caption Rewrites},
year = {2024},
isbn = {978-3-031-53301-3},
publisher = {Springer-Verlag},
address = {Berlin, Heidelberg},
url = {https://doi.org/10.1007/978-3-031-53302-0_3},
doi = {10.1007/978-3-031-53302-0_3},
booktitle = {MultiMedia Modeling: 30th International Conference, MMM 2024, Amsterdam, The Netherlands, January 29 – February 2, 2024, Proceedings, Part IV},
pages = {32–45},
numpages = {14},
location = {Amsterdam, The Netherlands}
}

@inproceedings{wu-etal-2024-autohallusion,
 address = {Miami, Florida, USA},
 author = {Wu, Xiyang  and
Guan, Tianrui  and
Li, Dianqi  and
Huang, Shuaiyi  and
Liu, Xiaoyu  and
Wang, Xijun  and
Xian, Ruiqi  and
Shrivastava, Abhinav  and
Huang, Furong  and
Boyd-Graber, Jordan Lee  and
Zhou, Tianyi  and
Manocha, Dinesh},
 booktitle = {Findings of the Association for Computational Linguistics: EMNLP 2024},
 doi = {10.18653/v1/2024.findings-emnlp.493},
 editor = {Al-Onaizan, Yaser  and
Bansal, Mohit  and
Chen, Yun-Nung},
 pages = {8395--8419},
 publisher = {Association for Computational Linguistics},
 title = {{A}uto{H}allusion: Automatic Generation of Hallucination Benchmarks for Vision-Language Models},
 url = {https://aclanthology.org/2024.findings-emnlp.493/},
 year = {2024}
}

@article{qiu2024longhalqa,
  title = {{LongHalQA}: Long-Context Hallucination Evaluation for MultiModal Large Language Models},
  author = {Qiu, Han and Huang, Jiaxing and Gao, Peng and Qi, Qin and Zhang, Xiaoqin and Shao, Ling and Lu, Shijian},
  journal = {arXiv preprint arXiv:2410.09962},
  year = {2024},
  url = {https://arxiv.org/abs/2410.09962}
}

@article{wang2024amber,
  title = {{AMBER}: An {LLM}-free Multi-dimensional Benchmark for {MLLMs} Hallucination Evaluation},
  author = {Wang, Junyang and Wang, Yuhang and Xu, Guohai and Zhang, Jing and Gu, Yukai and Jia, Haitao and Wang, Jiaqi and Xu, Haiyang and Yan, Ming and Zhang, Ji and Sang, Jitao},
  journal = {arXiv preprint arXiv:2311.07397},
  year = {2023},
  url = {https://arxiv.org/abs/2311.07397}
}

@INPROCEEDINGS{Villa_2025_CVPR,
  author={Villa, Andrés and Alcázar, Juan León and Soto, Alvaro and Ghanem, Bernard},
  booktitle={2025 IEEE/CVF Conference on Computer Vision and Pattern Recognition Workshops (CVPRW)}, 
  title={Behind the Magic, MERLIM: Multi-Modal Evaluation Benchmark for Large Image-Language Models}, 
  year={2025},
  volume={},
  number={},
  pages={492-502},
  doi={10.1109/CVPRW67362.2025.00054}}

@inproceedings{rohrbach-etal-2018-object,
 address = {Brussels, Belgium},
 author = {Rohrbach, Anna  and
Hendricks, Lisa Anne  and
Burns, Kaylee  and
Darrell, Trevor  and
Saenko, Kate},
 booktitle = {Proceedings of the 2018 Conference on Empirical Methods in Natural Language Processing},
 doi = {10.18653/v1/D18-1437},
 editor = {Riloff, Ellen  and
Chiang, David  and
Hockenmaier, Julia  and
Tsujii, Jun{'}ichi},
 pages = {4035--4045},
 publisher = {Association for Computational Linguistics},
 title = {Object Hallucination in Image Captioning},
 url = {https://aclanthology.org/D18-1437},
 year = {2018}
}

@inproceedings{ben-kish-etal-2024-mitigating,
 address = {Miami, Florida, USA},
 author = {Ben-Kish, Assaf  and
Yanuka, Moran  and
Alper, Morris  and
Giryes, Raja  and
Averbuch-Elor, Hadar},
 booktitle = {Proceedings of the 2024 Conference on Empirical Methods in Natural Language Processing},
 doi = {10.18653/v1/2024.emnlp-main.1263},
 editor = {Al-Onaizan, Yaser  and
Bansal, Mohit  and
Chen, Yun-Nung},
 pages = {22680--22698},
 publisher = {Association for Computational Linguistics},
 title = {Mitigating Open-Vocabulary Caption Hallucinations},
 url = {https://aclanthology.org/2024.emnlp-main.1263/},
 year = {2024}
}

@article{zhai2024halle,
  title = {{HallE-Control}: Controlling Object Hallucination in Large Multimodal Models},
  author = {Zhai, Bohan and Yang, Shijia and Xu, Chenfeng and Shen, Sheng and Keutzer, Kurt and Li, Chunyuan and Li, Manling},
  journal = {arXiv preprint arXiv:2310.01779},
  year = {2023},
  url = {https://arxiv.org/abs/2310.01779}
}

@inproceedings{lovenia-etal-2024-negative,
 address = {Bangkok, Thailand},
 author = {Lovenia, Holy  and
Dai, Wenliang  and
Cahyawijaya, Samuel  and
Ji, Ziwei  and
Fung, Pascale},
 booktitle = {Proceedings of the 3rd Workshop on Advances in Language and Vision Research (ALVR)},
 doi = {10.18653/v1/2024.alvr-1.4},
 editor = {Gu, Jing  and
Fu, Tsu-Jui (Ray)  and
Hudson, Drew  and
Celikyilmaz, Asli  and
Wang, William},
 pages = {37--58},
 publisher = {Association for Computational Linguistics},
 title = {Negative Object Presence Evaluation ({NOPE}) to Measure Object Hallucination in Vision-Language Models},
 url = {https://aclanthology.org/2024.alvr-1.4/},
 year = {2024}
}

@inproceedings{mscoco,
 address = {Cham},
 author = {Lin, Tsung-Yi
and Maire, Michael
and Belongie, Serge
and Hays, James
and Perona, Pietro
and Ramanan, Deva
and Doll{\'a}r, Piotr
and Zitnick, C. Lawrence},
 booktitle = {Computer Vision -- ECCV 2014},
 editor = {Fleet, David
and Pajdla, Tomas
and Schiele, Bernt
and Tuytelaars, Tinne},
 isbn = {978-3-319-10602-1},
 pages = {740--755},
 publisher = {Springer International Publishing},
 title = {Microsoft COCO: Common Objects in Context},
 year = {2014}
}

@article{chen2026survey,
  title={A survey of multimodal hallucination evaluation and detection},
  author={Chen, Zhiyuan and Min, Yuecong and Zhang, Jie and Yan, Bei and Wang, Jiahao and Wang, Xiaozhen and Shan, Shiguang},
  journal={International Journal of Computer Vision},
  volume={134},
  number={3},
  pages={131},
  year={2026},
  publisher={Springer}
}

@inproceedings{hudson2019gqa,
  author    = {Drew A. Hudson and Christopher D. Manning},
  title     = {{GQA}: A New Dataset for Real-World Visual Reasoning and Compositional Question Answering},
  booktitle = {Proceedings of the IEEE/CVF Conference on Computer Vision and Pattern Recognition (CVPR)},
  pages     = {6700--6709},
  year      = {2019},
  doi       = {10.1109/CVPR.2019.00686}
}

@inproceedings{balanced_vqa_v2,
  author    = {Yash Goyal and Tejas Khot and Douglas Summers-Stay and Dhruv Batra and Devi Parikh},
  title     = {Making the {V} in {VQA} Matter: Elevating the Role of Image Understanding in Visual Question Answering},
  booktitle = {Proceedings of the IEEE Conference on Computer Vision and Pattern Recognition (CVPR)},
  pages     = {6904--6913},
  year      = {2017},
  doi       = {10.1109/CVPR.2017.670}
}

@inproceedings{alwajih-etal-2024-dallah,
    title = "Dallah: A Dialect-Aware Multimodal Large Language Model for {A}rabic",
    author = "Alwajih, Fakhraddin  and
      Bhatia, Gagan  and
      Abdul-Mageed, Muhammad",
    editor = "Habash, Nizar  and
      Bouamor, Houda  and
      Eskander, Ramy  and
      Tomeh, Nadi  and
      Abu Farha, Ibrahim  and
      Abdelali, Ahmed  and
      Touileb, Samia  and
      Hamed, Injy  and
      Onaizan, Yaser  and
      Alhafni, Bashar  and
      Antoun, Wissam  and
      Khalifa, Salam  and
      Haddad, Hatem  and
      Zitouni, Imed  and
      AlKhamissi, Badr  and
      Almatham, Rawan  and
      Mrini, Khalil",
    booktitle = "Proceedings of the Second Arabic Natural Language Processing Conference",
    month = aug,
    year = "2024",
    address = "Bangkok, Thailand",
    publisher = "Association for Computational Linguistics",
    url = "https://aclanthology.org/2024.arabicnlp-1.27/",
    doi = "10.18653/v1/2024.arabicnlp-1.27",
    pages = "320--336"
}

@inproceedings{nayak-etal-2024-benchmarking,
  title = {Benchmarking Vision Language Models for Cultural Understanding},
  author = {Nayak, Shravan and Jain, Kanishk and Awal, Rabiul and Reddy, Siva and Steenkiste, Sjoerd Van and Hendricks, Lisa Anne and Stanczak, Karolina and Agrawal, Aishwarya},
  editor = {Al-Onaizan, Yaser and Bansal, Mohit and Chen, Yun-Nung},
  booktitle = {Proceedings of the 2024 Conference on Empirical Methods in Natural Language Processing},
  month = nov,
  year = {2024},
  address = {Miami, Florida, USA},
  publisher = {Association for Computational Linguistics},
  pages = {5769--5790},
  doi = {10.18653/v1/2024.emnlp-main.329},
  url = {https://aclanthology.org/2024.emnlp-main.329/}
}

@inproceedings{pfeiffer-etal-2022-xgqa,
  title = {x{GQA}: Cross-Lingual Visual Question Answering},
  author = {Pfeiffer, Jonas and Geigle, Gregor and Kamath, Aishwarya and Steitz, Jan-Martin O. and Roth, Stefan and Vuli{\'c}, Ivan and Gurevych, Iryna},
  editor = {Muresan, Smaranda and Nakov, Preslav and Villavicencio, Aline},
  booktitle = {Findings of the Association for Computational Linguistics: ACL 2022},
  month = may,
  year = {2022},
  address = {Dublin, Ireland},
  publisher = {Association for Computational Linguistics},
  pages = {2497--2511},
  doi = {10.18653/v1/2022.findings-acl.196},
  url = {https://aclanthology.org/2022.findings-acl.196/}
}

@inproceedings{urailertprasert-etal-2024-sea,
  title = {{SEA}-{VQA}: Southeast Asian Cultural Context Dataset For Visual Question Answering},
  author = {Urailertprasert, Norawit and Limkonchotiwat, Peerat and Suwajanakorn, Supasorn and Nutanong, Sarana},
  editor = {Gu, Jing and Fu, Tsu-Jui (Ray) and Hudson, Drew and Celikyilmaz, Asli and Wang, William},
  booktitle = {Proceedings of the 3rd Workshop on Advances in Language and Vision Research (ALVR)},
  month = aug,
  year = {2024},
  address = {Bangkok, Thailand},
  publisher = {Association for Computational Linguistics},
  pages = {173--185},
  doi = {10.18653/v1/2024.alvr-1.15},
  url = {https://aclanthology.org/2024.alvr-1.15/}
}

@inproceedings{ghaboura-etal-2025-camel,
  title = {{CAMEL}-Bench: A Comprehensive {A}rabic {LMM} Benchmark},
  author = {Ghaboura, Sara and Heakl, Ahmed and Thawakar, Omkar and Alharthi, Ali Husain Salem Abdulla and Riahi, Ines and Radman, Abduljalil and Laaksonen, Jorma and Khan, Fahad Shahbaz and Khan, Salman and Anwer, Rao Muhammad},
  editor = {Chiruzzo, Luis and Ritter, Alan and Wang, Lu},
  booktitle = {Findings of the Association for Computational Linguistics: NAACL 2025},
  month = apr,
  year = {2025},
  address = {Albuquerque, New Mexico},
  publisher = {Association for Computational Linguistics},
  pages = {1970--1980},
  doi = {10.18653/v1/2025.findings-naacl.105},
  url = {https://aclanthology.org/2025.findings-naacl.105/},
  isbn = {979-8-89176-195-7}
}

@inbook{zar2005,
author = {Zar, Jerrold H.},
publisher = {John Wiley \& Sons, Ltd},
isbn = {9780470011812},
title = {Spearman Rank Correlation},
booktitle = {Encyclopedia of Biostatistics},
chapter = {},
pages = {},
doi = {https://doi.org/10.1002/0470011815.b2a15150},
url = {https://onlinelibrary.wiley.com/doi/abs/10.1002/0470011815.b2a15150},
eprint = {https://onlinelibrary.wiley.com/doi/pdf/10.1002/0470011815.b2a15150},
year = {2005}
}

@inproceedings{alkulaib-etal-2026-datalabku,
  author    = {AlKulaib, Lulwah and Jlidi, Ahmed},
  title     = {{DATALabKU} at {ImageEval} 2026 Shared Tasks: Metric-Aware Cultural Evaluation for {CRAI-Bench}},
  booktitle = {Proceedings of the Fourth Arabic Natural Language Processing Conference: Shared Tasks},
  month     = oct,
  year      = {2026},
  address   = {Budapest, Hungary},
  publisher = {Association for Computational Linguistics}
}

@inproceedings{yamin-etal-2026-dynamos,
    author    = {Yamin, Muhammad  and
                 Wahid, Sheikh Abdul  and
                 Arshad, Ali},
    title     = "{Dynamos} at {ImageEval} 2026 Shared Tasks: Metric-Aligned Selection and
                 Cross-Family Ensembling for Vision-Language Hallucination Detection in
                 English and Arabic",
    booktitle = {Proceedings of the Fourth Arabic Natural Language Processing Conference: Shared Tasks},
    month     = oct,
    year      = {2026},
    address   = {Budapest, Hungary},
    publisher = {Association for Computational Linguistics}
}

@inproceedings{AlDahoul-etal-2026-NYUAD,
    author = {AlDahoul, Nouar and
              Zaki, Yasir},
    title = "{NYUAD} at {ImageEval} Shared Tasks: A Multimodal LLM for Audio-Visual Question Answering and Hallucination Detection in English and Modern Standard Arabic",
    booktitle = {Proceedings of the Fourth Arabic Natural Language Processing Conference: Shared Tasks},
    address = {Budapest, Hungary},
    month = oct,
    year = {2026},
    publisher = {Association for Computational Linguistics}
}

@inproceedings{zaman-etal-2026-nile-nexus,
    author    = {Zaman, Sumaiya  and
                 Rishta, Miftahul Jannat},
    title     = "{Nile Nexus} at {ImageEval} 2026 Shared Tasks: Single-Choice Reformulation and Confidence-Calibrated Refinement for Culturally Grounded Hallucination Detection",
    booktitle = {Proceedings of the Fourth Arabic Natural Language Processing Conference: Shared Tasks},
    month     = oct,
    year      = {2026},
    address   = {Budapest, Hungary},
    publisher = {Association for Computational Linguistics}
}

@inproceedings{ma-etal-2026-falcons,
    author = {Ma, Pan and Liu, Junlin and Liu, Jun},
    title = {{Team Falcons} at {ImageEval} 2026 Shared Tasks: Structure-Constrained {QLoRA} Fine-Tuning and Dual-Model Consensus Verification},
    booktitle = {Proceedings of the Fourth Arabic Natural Language Processing Conference: Shared Tasks},
    month = oct,
    year = {2026},
    address = {Budapest, Hungary},
    publisher = {Association for Computational Linguistics}
  }

@inproceedings{mohamed-etal-2026-ahmedayman,
    author    = {Mohamed, Ahmed Ayman Ahmed Ezzat},
    title     = "{Ahmed Ayman} at {ImageEval} 2026 Shared Tasks: Measurement-First Systems for Spoken {VQA} and Cultural Accuracy Evaluation",
    booktitle = {Proceedings of the Fourth Arabic Natural Language Processing Conference: Shared Tasks},
    month     = oct,
    year      = {2026},
    address   = {Budapest, Hungary},
    publisher = {Association for Computational Linguistics}
}

@inproceedings{sheikh-2026-mdfaisal,
    author    = {Sheikh, Md. Faisal},
    title     = "{md\_faisal} at {ImageEval} 2026 Shared Tasks: Forced-Choice Reformulation for Hallucination Detection and a Version-Prior Baseline for Cultural Relevance",
    booktitle = {Proceedings of the Fourth Arabic Natural Language Processing Conference: Shared Tasks},
    month     = oct,
    year      = {2026},
    address   = {Budapest, Hungary},
    publisher = {Association for Computational Linguistics}
}

@inproceedings{hoque-etal-2026-team-tokenizers,
    author    = {Hoque, Syed Mohaiminul  and
                 Hossain, Md Sakhawat},
    title     = "{Team Tokenizers} at {ImageEval} 2026 Shared Tasks: Answer-First Contrastive Grounding with QLoRA",
    booktitle = {Proceedings of the Fourth Arabic Natural Language Processing Conference: Shared Tasks},
    month     = oct,
    year      = {2026},
    address   = {Budapest, Hungary},
    publisher = {Association for Computational Linguistics}
}

@inproceedings{pasa-etal-2026-alkhder,
    author    = {Pasa, Maher and
                 Hanini, Maria and
                 Najjar, Amro and
                 Alkhder, Hasan},
    title     = "{alkhder} at {ImageEval} 2026 Shared Tasks: Constrained Selection and Rotation-Averaged Logit Scoring for Multimodal Hallucination Detection",
    booktitle = {Proceedings of the Fourth Arabic Natural Language Processing Conference: Shared Tasks},
    month     = oct,
    year      = {2026},
    address   = {Budapest, Hungary},
    publisher = {Association for Computational Linguistics}
}

@inproceedings{semon-etal-2026-cuet-inferx,
    author    = {Semon, Md. Ashraful Islam  and
                 Islam, Jihadul},
    title     = "{CUET\_InferX} at {ImageEval} 2026 Shared Tasks: Audio-Visual {LoRA} Adaptation of {Qwen2.5-Omni} for Culturally-Grounded Spoken {VQA}",
    booktitle = {Proceedings of the Fourth Arabic Natural Language Processing Conference: Shared Tasks},
    month     = oct,
    year      = {2026},
    address   = {Budapest, Hungary},
    publisher = {Association for Computational Linguistics}
}

@inproceedings{younis-etal-2026-ahmedyounis,
    author    = {Younis, Ahmed},
    title     = "{Ahmed Younis} at {ImageEval} 2026 Shared Tasks: The Judge Needs a Prior for Cultural Image Evaluation",
    booktitle = {Proceedings of the Fourth Arabic Natural Language Processing Conference: Shared Tasks},
    month     = oct,
    year      = {2026},
    address   = {Budapest, Hungary},
    publisher = {Association for Computational Linguistics}
}

@inproceedings{hassan-etal-2026-digilians,
  author = {Hassan, Ahmed Eid and
            Abd El-Maguid, Mohamed Eid and
            Rashad, Ahmed Hossam and
            Mahmoud, Nour Eldeen Hossam and
            Mohamed, Mohamed Hamdy and
            Fawzy, Abdelrhman Mahmoud},
  title = {Digilians at {ImageEval} 2026 Task 1a: Modular Speech Parsing and Joint Multimodal Reasoning for Arabic Spoken VQA},
  booktitle = {Proceedings of the Fourth Arabic Natural Language Processing Conference: Shared Tasks},
  month = oct,
  year = {2026},
  address = {Budapest, Hungary},
  publisher = {Association for Computational Linguistics}
}

\appendix
\section{Appendix}

\subsection{Development-Phase Results}
\label{app:devtest}
Tables~\ref{tab:dev-1a}, \ref{tab:dev-1b}, and \ref{tab:dev-2} report the development-phase results (each team's latest submission). For Task~1, we evaluated submissions on the dev-test split, using synthetic TTS audio for Task~1a; for Task~2, we used the development split. 

The results show several patterns. For Task~1a, English systems generally outperform their MSA counterparts, with a wider performance spread in MSA. Task~1b shows strong performance among the leading systems in both languages, although MSA remains more challenging overall. For Task~2, both participating systems substantially outperform the GPT-4o baseline. Overall, the development results indicate stronger performance in English for spoken VQA and consistent improvements over the released baselines across all tasks.


\begin{table}[!tbh]
\centering
\small
\setlength{\tabcolsep}{3.5pt}
\scalebox{0.9}{
\begin{tabular}{@{}c l ccc@{}}
\toprule
\textbf{\#} & \textbf{Team} & \textbf{Acc.} & \textbf{B-Acc.} & \textbf{M-F1} \\
\midrule
\multicolumn{5}{@{}l}{\textit{English}} \\
\midrule
1 & NYUAD & 0.974 & 0.974 & 0.974 \\
2 & romey101 & 0.972 & 0.972 & 0.972 \\
3 & CUET\_InferX & 0.916 & 0.916 & 0.916 \\
-- & \textit{Qwen2.5-Omni-3B (baseline)} & 0.664 & 0.666 & 0.661 \\
\addlinespace[3pt]
\multicolumn{5}{@{}l}{\textit{MSA}} \\
\midrule
1 & NYUAD & 0.920 & 0.919 & 0.921 \\
2 & Digilians & 0.784 & 0.780 & 0.782 \\
3 & md\_faisal & 0.398 & 0.410 & 0.344 \\
-- & \textit{Qwen2.5-Omni-3B (baseline)} & 0.398 & 0.410 & 0.344 \\
\bottomrule
\end{tabular}
}
\vspace{-0.3cm}
\caption{Task 1a development-phase results on the devtest split
(synthetic TTS audio). Qwen2.5-Omni-3B is the official baseline.}
\label{tab:dev-1a}
\vspace{-0.3cm}
\end{table}

\begin{table}[!tbh]
\centering
\small
\setlength{\tabcolsep}{3.5pt}
\scalebox{0.9}{
\begin{tabular}{@{}c l ccc@{}}
\toprule
\textbf{\#} & \textbf{Team} & \textbf{CI}$\downarrow$ &
\textbf{Acc.}$_{Q^+}$ & \textbf{Acc.}$_{Q^-}$ \\
\midrule
\multicolumn{5}{@{}l}{\textit{English}} \\
\midrule
1 & Team Tokenizers & 0.028 & 0.972 & 0.986 \\
2 & NYUAD & 0.028 & 0.972 & 0.986 \\
3 & Team Falcons & 0.030 & 0.970 & 0.985 \\
4 & nawwad & 0.036 & 0.964 & 0.982 \\
5 & Nile Nexus & 0.050 & 0.950 & 0.975 \\
6 & ameya & 0.284 & 0.716 & 0.858 \\
7 & md\_faisal & 0.313 & 0.940 & 0.839 \\
-- & \textit{Qwen2.5-VL-3B (baseline)} & 0.313 & 0.940 & 0.839 \\
\addlinespace[3pt]
\multicolumn{5}{@{}l}{\textit{MSA}} \\
\midrule
1 & NYUAD & 0.036 & 0.964 & 0.982 \\
2 & NAMAA Community & 0.046 & 0.954 & 0.977 \\
3 & lettycat & 0.082 & 0.918 & 0.959 \\
4 & md\_faisal & 0.490 & 0.834 & 0.773 \\
-- & \textit{Qwen2.5-VL-3B (baseline)} & 0.490 & 0.834 & 0.773 \\
\bottomrule
\end{tabular}
}
\vspace{-0.2cm}
\caption{Task 1b development-phase results on the devtest split,
ranked by contrastive instability (CI, lower is better). Qwen2.5-VL-3B
is the official baseline.}
\label{tab:dev-1b}
\vspace{-0.3cm}
\end{table}

\begin{table}[]
\centering
\small
\setlength{\tabcolsep}{3.5pt}
\scalebox{0.9}{
\begin{tabular}{@{}c l cc@{}}
\toprule
\textbf{\#} & \textbf{Team} & $\boldsymbol\rho\uparrow$ & \textbf{MAE}$\downarrow$ \\
\midrule
1 & Ahmed Younis & 0.708 & 0.188 \\
2 & md\_faisal & 0.641 & 0.204 \\
-- & \textit{GPT-4o (baseline)} & 0.215 & 0.367 \\
\bottomrule
\end{tabular}
}
\vspace{-0.2cm}
\caption{Task 2 development-phase results on the dev split, ranked by
Spearman correlation of the composite score. The GPT-4o judge is the
official baseline.}
\label{tab:dev-2}
\vspace{-0.3cm}
\end{table}

\end{document}